\documentclass[letterpaper, 10 pt, conference]{ieeeconf}  

\IEEEoverridecommandlockouts                              

\usepackage{graphicx}
\usepackage{caption}
\graphicspath{
    {figures/}
}

\usepackage{amsmath,amssymb,enumerate}

\usepackage{amsthm}
\usepackage{setspace}
\usepackage{booktabs}
\usepackage[usenames,dvipsnames,svgnames,table]{xcolor}
\usepackage{mathtools}
\usepackage{algorithm, algorithmicx, algpseudocode}
\usepackage{blindtext}
\usepackage{gensymb}
\usepackage{xparse}
\usepackage{lipsum}
\usepackage{mathrsfs}
\usepackage[mathscr]{euscript}
\usepackage{times}
\usepackage{cite} 
\usepackage{multicol}
\definecolor{darkgreen}{rgb}{0,0.6,0}
\usepackage[bookmarks=true,colorlinks=true,pdfpagemode=UseNone,citecolor=darkgreen,linkcolor=black,urlcolor=BrickRed,pagebackref]{hyperref}
\usepackage[caption=false,font=footnotesize]{subfig}
\usepackage{amsfonts}
\usepackage[utf8]{inputenc}
\usepackage[T1]{fontenc}
\usepackage{textcomp}
\usepackage{arydshln}
\usepackage{balance}

\newtheorem{problem}{Problem}

\newtheorem{remark}{Remark}

\renewcommand{\thefigure}{\arabic{figure}}

\definecolor{note}{rgb}{0.1,0.1,1}
\definecolor{rephase}{rgb}{0.15,0.7,0.15}
\definecolor{bag}{rgb}{0.6,0.6,0.2}

\makeatletter
\renewcommand*\env@matrix[1][c]{\hskip -\arraycolsep
  \let\@ifnextchar\new@ifnextchar
  \array{*\c@MaxMatrixCols #1}}
\makeatother

\newcommand{\transpose}{\mathsf{T}}

\makeatletter
\newcommand{\mathleft}{\@fleqntrue\@mathmargin0pt}
\newcommand{\mathcenter}{\@fleqnfalse}
\makeatother

\definecolor{orange}{RGB}{255,127,0}

\title{\LARGE \bf An Error-State Model Predictive Control on Connected Matrix Lie~Groups for Legged Robot Control} 

\author{Sangli Teng, Dianhao Chen, William Clark, and Maani Ghaffari%
\thanks{Toyota Research Institute provided funds to support this work. Funding for M. Ghaffari was in part provided by NSF Award No. 2118818. This work was also supported by MIT Biomimetic Robotics Lab and NAVER LABS. W. Clark was supported by NSF grant DMS-1645643.}
\thanks{S.~Teng, D.~Chen, and M.~Ghaffari are with the University of Michigan, Ann Arbor, MI 48109, USA. {\tt\small\{sanglit,chendh,maanigj\} @umich.edu}}%
\thanks{W.~Clark is with the Department of Mathematics, Cornell University, Ithaca, NY. {\tt\small wac76@cornell.edu}}
}

\begin{document}

\maketitle
\thispagestyle{empty}
\pagestyle{empty}

\setlength{\belowdisplayskip}{2pt}
\setlength{\textfloatsep}{4pt}	

\begin{abstract}
This paper reports on a new error-state Model Predictive Control (MPC) approach to connected matrix Lie groups for robot control. The linearized tracking error dynamics and the linearized equations of motion are derived in the Lie algebra. Moreover, given an initial condition, the linearized tracking error dynamics and equations of motion are globally valid and evolve independently of the system trajectory. By exploiting the symmetry of the problem, the proposed approach shows faster convergence of rotation and position simultaneously than the state-of-the-art geometric variational MPC based on variational-based linearization. Numerical simulation on tracking control of a fully-actuated 3D rigid body dynamics confirms the benefits of the proposed approach compared to the baselines. Furthermore, the proposed MPC is also verified in pose control and locomotion experiments on a quadrupedal robot MIT Mini Cheetah.

\end{abstract} 

\IEEEpeerreviewmaketitle

\section{Introduction}
The geometry of the configuration space of a robotics system can naturally be modeled using matrix Lie (continuous) groups~\cite{bloch2015nonholonomic,lynch2017modern}. Moreover, Lie group techniques have been successfully used to study the symmetry structures of control and observer systems~\cite{grizzle1985structure,bonnabel2009non,barrau2017invariant,chetverikov2022orbital}. For example, the unmanned aerial vehicles and the centroidal dynamics of legged robots can be approximated by a single rigid body, whose motion is on $\mathrm{SE}(3)$. The $\mathrm{SE}(3)$ manifold is different from $\mathbb{R}^n$ Euclidean space, where most controllers are designed and applied. Although one can represent the orientation of the robot by the rotation matrix, many applications use the Euler angles \cite{park2017high}, or quaternions \cite{bloesch2013state}. However, the Euler angles are known for singularities in some configurations \cite{Euler-survey}, and quaternions have ambiguities in representing the attitudes \cite{quaternion}. Geometric Model Predictive Control (MPC) \cite{yanran-NMPC-variational,vblMPC} has been proposed to address these challenges. However, these approaches do not exploit the existing symmetry of pose control problem on $\mathrm{SE}(3)$ Lie group or assume the current system trajectory is sufficiently close to the desired trajectory. This assumption might not be satisfied in practice. 

Geometric control techniques on manifolds attempt to overcome the challenge in control by extracting the intrinsic property of the mechanical system \cite{bullo2019geometric, bullo1999tracking}. Locally exponentially stable tracking controllers for quadrotors are proposed in \cite{lee2010geometric,lee2011geometric} using the compatible error \cite{bullo2019geometric} on $\mathrm{SO}(3)$ to overcome the problem caused by Euler angles and quaternions. This research provides us with an abundant reference to formulate the error dynamics on manifolds. 

\begin{figure}[t]
    \centering
    \includegraphics[width=0.9\columnwidth]{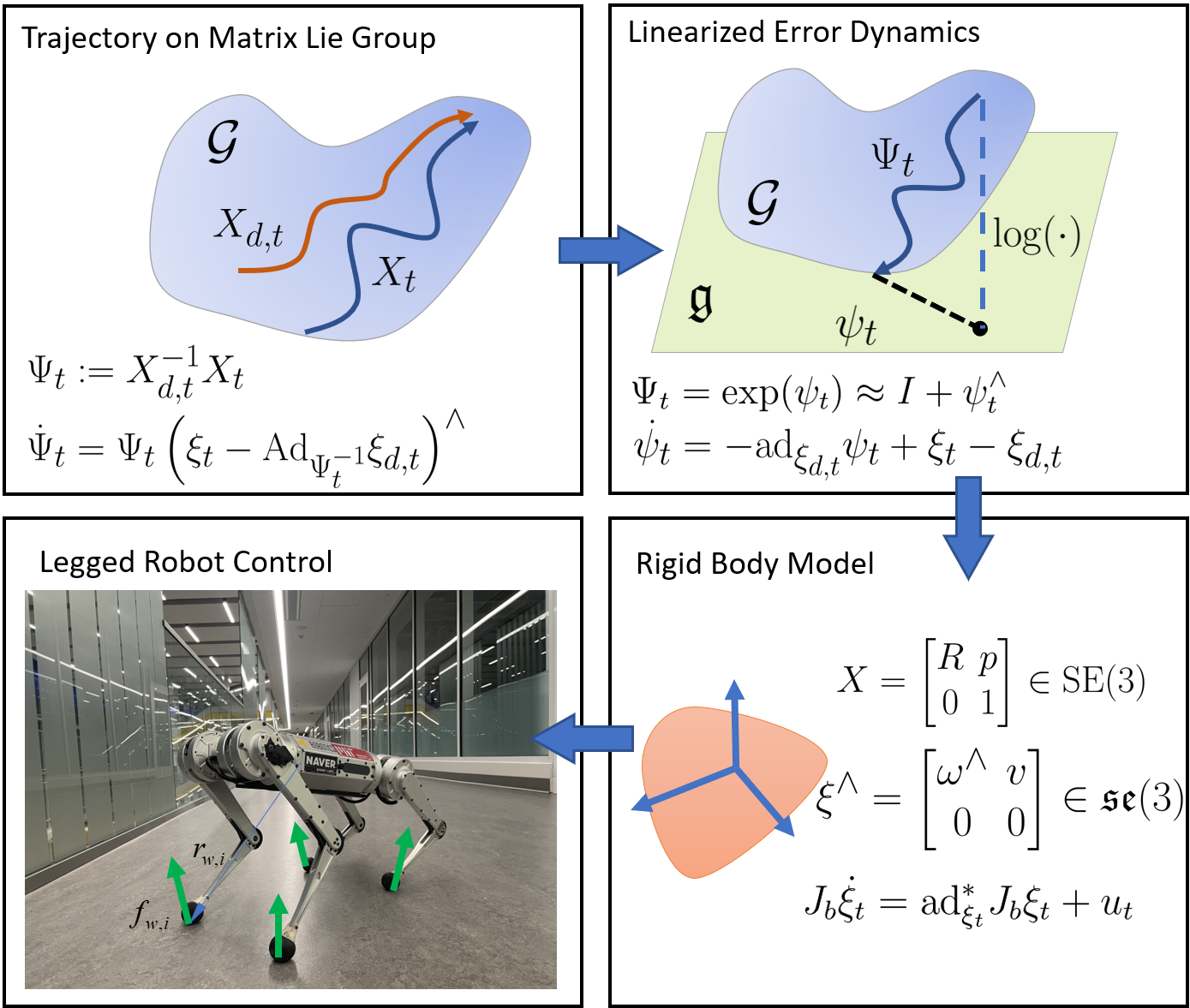}
    \caption{The proposed error-state MPC framework. The tracking error is defined on a matrix Lie group and linearized in the Lie algebra. A convex MPC algorithm is derived via the linearized dynamics for tracking control. The proposed algorithm is applied to a single rigid body system and verified on a quadrupedal robot MIT Mini Cheetah.} 
    \label{fig:first_fig}
\end{figure}

Nonlinear MPC (NMPC) has been proposed for rigid body systems tracking control. An MPC for discrete dynamics on $\mathrm{SO}(3)$ is developed in \cite{Ilya-SO3} for spacecraft attitude control. Using the matrix logarithm map, the controller can generate discontinuous control law that can achieve globally asymptotic stability. Similar techniques have also been applied to $\mathrm{SE}(3)$ in \cite{Ilya-SE3}. To preserve the energy of the system, an implicit dynamics equation obtained by the Lie group Variational Integrator \cite{leeLGVI} needs to be solved in \cite{Ilya-SE3, Ilya-SO3}. A nonlinear least-squares problem on $\mathrm{SO}(3)$ has been proposed in \cite{park-NMPC-euler-lin} for control of a legged robot approximated as a single rigid body. The Jacobian and Hessian at the tangent space of the orientation manifold are derived to approximate the least-squares problem. Differential Dynamics Programming (DDP) is also applied to the Lie group, which can be potentially applied in a receding horizon manner as MPC for optimal control \cite{lie-ddp} and state estimation \cite{kobilarov2015differential}. Moreover, factor graphs have been applied to estimation and control problems on Lie groups~\cite{ta2014factor}. 

Convex MPC has also been proposed for tracking control of rigid body dynamics. The Euler angle-based convex MPC \cite{CheetahCMPC} has been proposed for locomotion planning on the quadrupedal robot, which needs to assume zero pitch and roll angle. A local control law has been proposed in \cite{Ilya-SE3,Ilya-SO3}, where the linearized dynamics are defined by a local diffeomorphism from the $\mathrm{SE}(3)$ manifold to $\mathbb{R}^n$ space. However, such a diffeomorphism is not unique and too abstract for controller design. The Variational Based Linearization (VBL) technique \cite{Wu-VBL} is applied to generate linearized dynamics of the single rigid body around a given trajectory and applied to robot pose control \cite{Chig-VBL}. A VBL-based MPC is proposed in \cite{vblMPC} for locomotion on discrete terrain using a gait library. The result suggests that the VBL-based linearization can preserve the Lagrangian, thus making the system more stable. Other than linearizing at the reference trajectory, the work of \cite{yanran-NMPC-variational} linearizes the system at the current operating point to obtain the Quadratic Programming (QP) problem for tracking of legged robot trajectory. However, the linearized state matrix of \cite{yanran-NMPC-variational} depends on the orientation, which one can avoid by exploiting the symmetry of the system as done in this work. More recent work has been proposed to exploit the symmetry of rigid body dynamics. The work of~\cite{hampsey2022exploiting} studies the equivariant system on Lie groups, which induces a state-independent linearization scheme for quadrotors when taking the angular velocities as state inputs.

In this paper, we develop a geometric error-state MPC for tracking control of systems evolving on a Matrix Lie group, specifically, on $\mathrm{SE}(3)$ for the rigid body motion control. In particular, the main contributions of this work are as follows. 
\begin{enumerate}
    \item We derive the linearized tracking error dynamics and the linearized equations of motion in the Lie algebra (tangent space at the identity) that, given an initial condition, are globally valid and evolve independently of the system trajectory.
    \item We develop a convex MPC algorithm for the tracking control problem using the linearized error dynamics, which can be solved efficiently using QP solvers. 
    \item The proposed controller is validated via numerical simulations and in experiments on quadrupedal robot pose control and locomotion.
    \item Implementation of the proposed MPC is available for download at \url{https://github.com/UMich-CURLY/Error-State-MPC}. 
\end{enumerate}
The remainder of this paper is organized as follows. Section II provides the mathematical preliminaries and definitions used throughout the paper. Section III presents the error-state convex MPC. Numerical simulation and experiments are presented in Section IV and V, respectively. Discussions about the experiments are presented in Section VI. Section VII concludes the paper and discusses ideas for future studies.

\section{Problem Statement}
This section provides a brief overview of the necessary background used in the developed approach.
\subsection{Mathematical preliminary}
Let $\mathcal{G}$ be an $n$-dimensional matrix Lie group and $\mathfrak{g}$ its associate Lie algebra (hence, $\dim \mathfrak{g} = n$)~\cite{chirikjian2011stochastic,hall2015lie}. For convenience, we define the following isomorphism
\begin{equation}
    (\cdot)^\wedge:\mathbb{R}^n \rightarrow \mathfrak{g},
\end{equation}
that maps an element in the vector space $\mathbb{R}^n$ to the tangent space of the matrix Lie group at the identity. Then, for any $\phi \in \mathbb{R}^{n}$, we can define the Lie exponential map as
\begin{equation}
    \exp(\cdot):\mathbb{R}^{n} \rightarrow \mathcal{G},\ \ \exp(\phi)=\operatorname{exp_m}({\phi}^\wedge),
\end{equation}
where $\operatorname{exp_m}(\cdot)$ is the exponential of square matrices.
For every $X \in \mathcal{G}$, the adjoint action, $\mathrm{Ad}_{X}: \mathfrak{g}\rightarrow \mathfrak{g}$, is a Lie algebra isomorphism that enables change of frames 
\begin{equation}
    \mathrm{Ad}_{X}({\phi}^\wedge)= X{{\phi}^\wedge}X^{-1}.
\end{equation}
Its derivative at the identity gives rise to the adjoint map in Lie Algebra as
\begin{equation}
    \mathrm{ad}_{\phi}(\eta) = [{\phi}^\wedge, {\eta}^\wedge],
\end{equation}
where $\phi^\wedge, \eta^\wedge \in \mathfrak{g}$ and $[\cdot, \cdot]$ is the Lie bracket. 

Consider the motion of an object whose state space is a Lie group $\mathcal{G}$. We define a left-invariant Lagrangian $\mathcal{L}:\mathfrak{g}\to\mathbb{R}$ as
\begin{equation*}
    \mathcal{L}(\xi) = \frac{1}{2} \xi^\transpose J_b \xi,
\end{equation*}
where $\xi$ is the twist in the body frame, and $J_b$ is the generalized inertia matrix in the body fixed principal axes. We can then write the \emph{forced} Euler-Poincar\'{e} equations~\cite{Bloch1996}:
\begin{align}
\label{eq:forcedEP}
    J_b \dot{\xi}  = \mathrm{ad}^*_\xi J_b \xi + u, 
\end{align}
where $u \in \mathfrak{g}^*$ is the generalized control input force applied to the body fixed principal axes, $\mathrm{ad}^*$ is the coadjoint action, and $\mathfrak{g}^{*}$ is the cotangent space. Please see~\cite{bloch2015nonholonomic} for more background. 
\subsection{Rigid body dynamics}
Now consider a 3D rigid body in $\mathrm{SE}(3)$, the state of the robot can be represented by a rotation matrix $$R \in \mathrm{SO}(3) = \{R\in \mathbb{R}^{3 \times 3} \mid R^{\transpose}R = I_3, \det(R) = 1\},$$ and position $p \in \mathbb{R}^3$. We denote the identity matrix by $I$, and $I_3$ denotes the $3\times 3$ identity matrix. Then the homogeneous representation of an element in $\mathrm{SE}(3)$ is given by
\begin{equation}
    X = \begin{bmatrix}
    R & p \\ 0 & 1
    \end{bmatrix} \in \mathrm{SE}(3).
\end{equation}
We define the twist as the concatenation of linear velocity $v$ and angular velocity $\omega$ in body frame, i.e., $\xi := \begin{bmatrix} \omega \\ v \end{bmatrix} \in \mathbb{R}^6$, 
$\xi^{\wedge}=\begin{bmatrix} \omega^\wedge & v \\ 0 & 0 \end{bmatrix}\in\mathfrak{se}(3).$
The inertia matrix is defined as 
\begin{equation}\label{eq:inertia_matrix}
    J_b := \begin{bmatrix} I_b & 0 \\ 0 & mI_3 \end{bmatrix},
\end{equation}
where $I_b$ is the moment of inertia in the body frame, and $m$ is the body mass. 
The matrix representation of the adjoint map can be derived as
\begin{equation}
    \mathrm{Ad}_{X} = \begin{bmatrix}
    R & 0 \\ {p}^\wedge R & R
    \end{bmatrix},\quad X = \begin{bmatrix}
    R & p \\ 0 & 1
    \end{bmatrix}.
\end{equation}
Furthermore, the matrix representation of the adjoint in Lie Algebra is
\begin{equation}
    \mathrm{ad}_{\xi} = \begin{bmatrix}
    {\omega}^\wedge & 0 \\ 
    {v}^\wedge & {\omega}^\wedge
    \end{bmatrix}.
\end{equation}
Then the coadjoint map is 
\begin{equation}
    \mathrm{ad}^*_{\xi} = \mathrm{ad}_{\xi}^\transpose = -\begin{bmatrix}
    \omega^\wedge & v^\wedge \\ 
    0 & \omega^\wedge
    \end{bmatrix}.
\end{equation}
Finally, using \eqref{eq:forcedEP} combined with the reconstruction equation of $X \in \mathrm{SE}(3)$, i.e., \mbox{$\dot{X} = X\xi^\wedge$}, we arrive at the rigid body equation of motion
\begin{align}
\label{eq:rigid_body_dynamics}
    \nonumber J_b \dot{\xi}  + \begin{bmatrix}
    \omega^\wedge & v^\wedge \\ 0 & \omega^\wedge
    \end{bmatrix} J_b \xi = u, \\
    \begin{bmatrix}
    \dot{R} & \dot{p} \\ 0 & 0
    \end{bmatrix} = \begin{bmatrix}
    R & p \\ 0 & 1
    \end{bmatrix} \begin{bmatrix}
    {\omega}^\wedge & v \\ 0 & 0
    \end{bmatrix}.
\end{align}

\subsection{Tracking error dynamics}
Consider the trajectory on Lie group $\mathcal{G}$ , we define the desired trajectory as $X_{d,t} \in \mathcal{G}$ and the actual state as $X_t \in \mathcal{G}$, both as function of time $t$. Given the twists $\xi_t$ and desired twists $\xi_{d,t}$, we have
\begin{align*}
    \frac{d}{dt} X_t = X_t {\xi}^\wedge_t, \ \frac{d}{dt} X_{d, t} = X_{d,t} {\xi}^\wedge_{d,t}.
\end{align*}
Similar to the left or right error defined in \cite{bullo1999tracking}, we define the error between $X_t^d$ and $X_t$ as
\begin{equation}
\label{eq:X_err}
    \Psi_t = X_{d,t}^{-1} X_t \in \mathcal{G}. 
\end{equation}
For the tracking problem, our goal is to drive the error from the initial condition $\Psi_0$ to the identity $I \in \mathcal{G}$. Taking derivative on both sides of~\eqref{eq:X_err}, we have
\begin{align*}
    \frac{d}{dt} \Psi_t &= \dot{\Psi}_t =\frac{d}{dt} (X_{d,t}^{-1}) X_t + X_{d,t}^{-1}\frac{d}{dt} X_t \\
    &=X_{d,t}^{-1}\frac{d}{dt} X_t-X_{d,t}^{-1}\frac{d}{dt}(X_{d,t})X_{d,t}^{-1} X_t\\
    &=X_{d,t}^{-1}X_t{\xi}^\wedge_t - X_{d,t}^{-1}X_{d,t} {\xi}^\wedge_{d,t}X_{d,t}^{-1}X_t \\
    &=\Psi_t{\xi}^\wedge_t-{\xi}^\wedge_{d,t}\Psi_t.
\end{align*}
Therefore, 
\begin{equation}
\label{eq:error_dynamics}
\begin{aligned}
\dot{\Psi}_t &= \Psi_t({\xi}_t^\wedge - \Psi_t^{-1}{\xi}_{d,t}^\wedge \Psi_t) =\Psi_t({\xi}_t - \mathrm{Ad}_{\Psi_t^{-1}}{\xi}_{d,t})^\wedge,
\end{aligned}
\end{equation}
where $\mathrm{Ad}_{\Psi_t^{-1}}$ describes the transport map in~\cite{bullo1999tracking} that enables the comparison of velocities from different reference frames.
\subsection{Tracking control problem}
Given the rigid body dynamics \eqref{eq:forcedEP} and tracking error dynamics \eqref{eq:error_dynamics}, we define the tracking control problem as follows.
\begin{problem}
\label{prob:nmpc}
Find $u_t \in \mathfrak{g}^*$ such that 
\begin{align*}
    \min_{u_t} \quad & N(\Psi_{t_f}, \xi_{t_f}) + \int_{0}^{t_f} L(\Psi_t, \xi_t, u_t) \ dt \\
    \text{s.t. } & \dot{\Psi}_t =\Psi_t( {\xi}_t - \mathrm{Ad}_{\Psi_t^{-1}}{\xi}_{d,t})^{\wedge} \\
    & \dot{\xi}_t  = J_b^{-1}\left(\mathrm{ad}^*_{\xi_t} J_b \xi_t + u_t\right) \\
    & u_t \in \mathcal{U}_t, \xi(0) = \xi_0, \Psi(0) = \Psi_0,
\end{align*}
\end{problem}
\noindent where $t_f$ is the final time, $N(\cdot)$ is the terminal cost, $L(\cdot)$ is the stage cost, and $\mathcal{U}_t$ is the set of feasible input at time $t$. 


\section{Error-State Convex MPC}

\subsection{System linearization}
Problem~\ref{prob:nmpc} is nonconvex and evolves on a matrix Lie group. To implement it in real-time efficiently, we linearize and vectorize it in the following.

Recall that we can map the error from the Lie Algebra to the group element by the group exponential map. We define ${\psi}^{\wedge}_t$ as an element of the Lie Algebra that corresponds to $\Psi_t$. Thus by the exponential map, we have $$\Psi_t=\exp(\psi_t),\ \Psi_t\in\mathcal{G},\ {\psi}^{\wedge}_t\in\mathfrak{g}.$$ Given the first-order approximation of the exponential map,
$$\Psi_t=\exp(\psi_t) \approx I + {\psi}^{\wedge}_t, $$
and a first-order approximation of the adjoint map $$\mathrm{Ad}_{\Psi_t} \approx \mathrm{Ad}_{I + {\psi}^{\wedge}_t},$$
we can linearize~\eqref{eq:error_dynamics} by dropping the second-order terms as
\begin{equation}
    \dot{\Psi}_t\approx(I + \dot{{\psi}}^{\wedge}_t) \approx (I + {\psi}^{\wedge}_t)({\xi}_t - \mathrm{Ad}_{(I - {\psi}_t^{\wedge})}{\xi}_{d,t})^{\wedge},
\end{equation}
\begin{equation}
\label{eq:se3_error_lin}
    \dot{\psi}_t = -\mathrm{ad}_{\xi_{d,t}}\psi_t + \xi_t - \xi_{d,t}.
\end{equation}
Equation~\eqref{eq:se3_error_lin} is the linearized velocity error in the Lie algebra.
\begin{remark}
The reason behind lifting the problem to the Lie algebra is that one can use the usual algebraic manipulations and differential equations knowledge to formulate the problem. This approach also enables us to use the existing QP solvers, as we will see in the following sections.
\end{remark}

The dynamics of $\xi_t$ is described by \eqref{eq:forcedEP}, which is nonlinear. To compute a locally linear approximation of the nonlinear term, we adopt the following series expansion around the operating point $\bar{\xi}$
\begin{equation}
\label{eq:forcedEP_lin_1}
    J_b\dot{\xi} \approx \mathrm{ad}^*_{\bar{\xi}} J_b\bar{\xi} + \frac{\partial \mathrm{ad}^*_\xi J_b\xi}{\partial{\xi}}|_{\bar{\xi}} (\xi - \bar{\xi}) + u .
\end{equation}
Thus, we have the linearized dynamics as
\begin{equation}
\label{eq:forcedEP_lin_2}
\dot{\xi} = H_t\xi + J_b^{-1}u + b_t,
\end{equation}
where $H_t$ and $b_t$ are as follows.
\begin{equation}
\begin{aligned}
    H&:=J_b^{-1}\mathrm{ad}_{\bar{\xi}}^{*}J_b + J_b^{-1} \begin{bmatrix}
    (I_b{\bar{\omega}})^{\wedge} & m{\bar{v}^{\wedge}} \\
    m{\bar{v}^{\wedge}} & 0
    \end{bmatrix}, \\
    b_t&:=-J_b^{-1}\begin{bmatrix}
    (I_b{\bar{\omega}})^{\wedge} & m{\bar{v}^{\wedge}} \\
    m{\bar{v}^{\wedge}} & 0
    \end{bmatrix}\bar{\xi}.
\end{aligned}
\end{equation}
Note that we obtained $H_t$ via the chain rule, and its compact form is attributed to those blocks with $(\cdot)^{\wedge}$ in the coadjoint map.
We define the system states as $x_t:=\begin{bmatrix} \psi_t \\ \xi_t \end{bmatrix}$. Then, the linearized dynamics becomes
\begin{equation}
\label{eq:ct_dynamics}
    \dot{x}_t = A_tx_t + B_t u_t + h_t,
\end{equation}
\noindent where
\begin{align*}
    A_t:=\begin{bmatrix}
    -\mathrm{ad}_{\xi_{d,t}} & I \\
    0 & H_t
    \end{bmatrix}, B_t:=\begin{bmatrix}
    0 \\
    J_b^{-1}
    \end{bmatrix}, h_t:=\begin{bmatrix}
    -\xi_{d,t},\\
    b_t
    \end{bmatrix}.
\end{align*}
\begin{remark}
The operating point $\bar{\xi}$, for computing $H$ and $b$, need not to be the reference trajectory $\xi_{d,t}$. In the following sections, we set the operating point at the current system states when the controller is applied, which exhibits higher stability as shown by \cite{yanran-NMPC-variational}.  
\end{remark}
\subsection{Cost function for tracking control}
In $\mathbb{R}^n$ Euclidean space, we could directly penalize the difference between the actual and desired velocities. However, on Lie groups, the velocity vectors at different locations on the manifold cannot be compared directly. Instead, we need a transport map that moves the velocity to the same reference frame. Therefore, our cost function is designed to regulate the tracking error $\psi_t$ and its derivative $\dot{\psi}_t$ rather than the difference between $\xi_{d,t}$ and $\xi_t$. 

Thus, our tracking error can be designed as $y_t:=\begin{bmatrix} \psi_t \\ \dot{\psi}_t \end{bmatrix}$. Then, $y_t$ can be expressed by
\begin{equation}
    \begin{aligned}
        y_t &= C_tx_t - d_t, \\
        C_t &:= \begin{bmatrix}
        I & 0 \\
        -\mathrm{ad}_{\xi_{d,t}} & I
        \end{bmatrix} , d_t = 
        \begin{bmatrix}
        0 \\ 
        \xi_{d,t}
        \end{bmatrix}.
    \end{aligned}
\end{equation}
Given some semi-positive definite matrices $P$, $Q$, and $R$, we can now write the cost function as
\begin{equation}
    N(y_{t_f}) = y_{t_f}^{\transpose}Py_{t_f}, \ L(y_{t}, u_t) = y_{t}^{\transpose}Qy_{t} + u_t^{\transpose}Ru_t .
\end{equation}

\subsection{The convex MPC problem}
Given the cost function provided in the last section, we derive the following linear quadratic tracking style problem in the finite-time horizon.
\begin{problem}
\label{prob:cmpc}
Find $u_t \in \mathfrak{g}^*$ such that
\begin{align*}
    \min_{u_t} \quad & N(y_{t_f}) + \int_{0}^{t_f} L(y_t, u_t) \ dt \\
    \text{s.t. } & \dot{x}_t = A_tx_t + B_tu_t + h_t \\
    & u_t \in \mathcal{U}_t, \xi(0) = \xi_0, \psi(0) = \psi_0.
\end{align*}
\end{problem}

Given the future twists $\xi_{d,t}$, initial error state $\psi_0$, and twist $\xi_0$, we can define all the matrices. By discretizing the system at time steps $\{t_k\}_{k=1}^N$ and applying the controller in a receding horizon manner, we can derive the MPC problem in discrete-time as follows.
\begin{problem}
\label{prob:dtcmpc}
Find $u_k \in \mathfrak{g}^*$ such that
\begin{align*}
    \min_{u_k} \quad & y_N^{\transpose}Py_N + \sum_{k=1}^{N-1}y_k^{\transpose}Qy_k + u_k^{\transpose}Ru_k \\
    \text{s.t. } & x_{k+1} = A_{k}x_k + B_ku_k + h_k \\
    & u_k \in \mathcal{U}_k, x(0) = x_0 \\
    & k = 0,1,\dots,N-1.
\end{align*}
\end{problem}
In Problem~\ref{prob:dtcmpc}, $A_k$, $B_k$, and $h_k$ can be obtained by zero-order hold or Euler first-order integration. Problem~\ref{prob:dtcmpc} is a QP problem that can be solved efficiently, e.g., using OSQP~\cite{osqp}. 

\begin{remark}
In the presented experiments and simulations, for simplicity, we apply the Euler first-order integration, such that 
$$A_k = I + A_{t_k} \Delta t, \ B_k = B_{t_k}\Delta t, \ h_k = h_{t_k} \Delta t.$$
The comparison of different integration techniques in the context of the proposed approach is an interesting future research direction. 
\end{remark}

\begin{figure*}
    \centering
    \includegraphics[width=1.99\columnwidth]{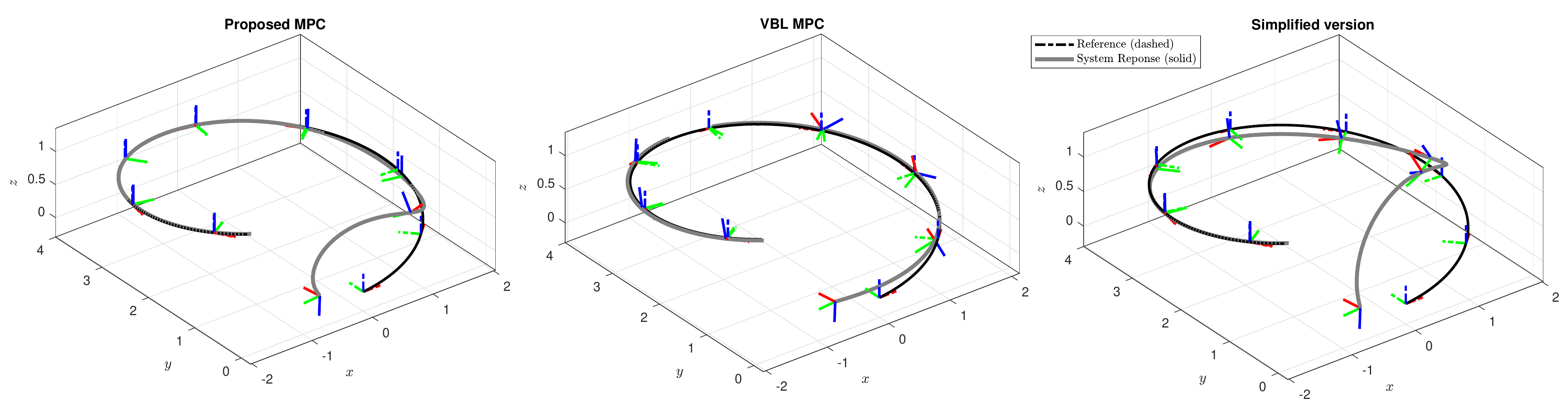}
    \caption{Simulation of a 3D rigid body tracking a spiral curve. The red, green, and blue denote the robot $x,y$, and $z$ axis, respectively. The dashed line is the reference trajectory, and the solid line indicates the system response with the corresponding controller. We chose one initial condition with a large orientation error in this figure. We can see that the orientation fast converges to the ground reference using our controller. The VBL MPC is much slower. As our controller deals with the position in the body frame, the convergence of position is not as fast as the VBL MPC. The simplified version has even slower position error convergence.}
    \label{fig:sim_spiral}
\end{figure*}

\section{Numerical Simulations}

\begin{figure}[t]
    \centering
    \includegraphics[width=1\columnwidth]{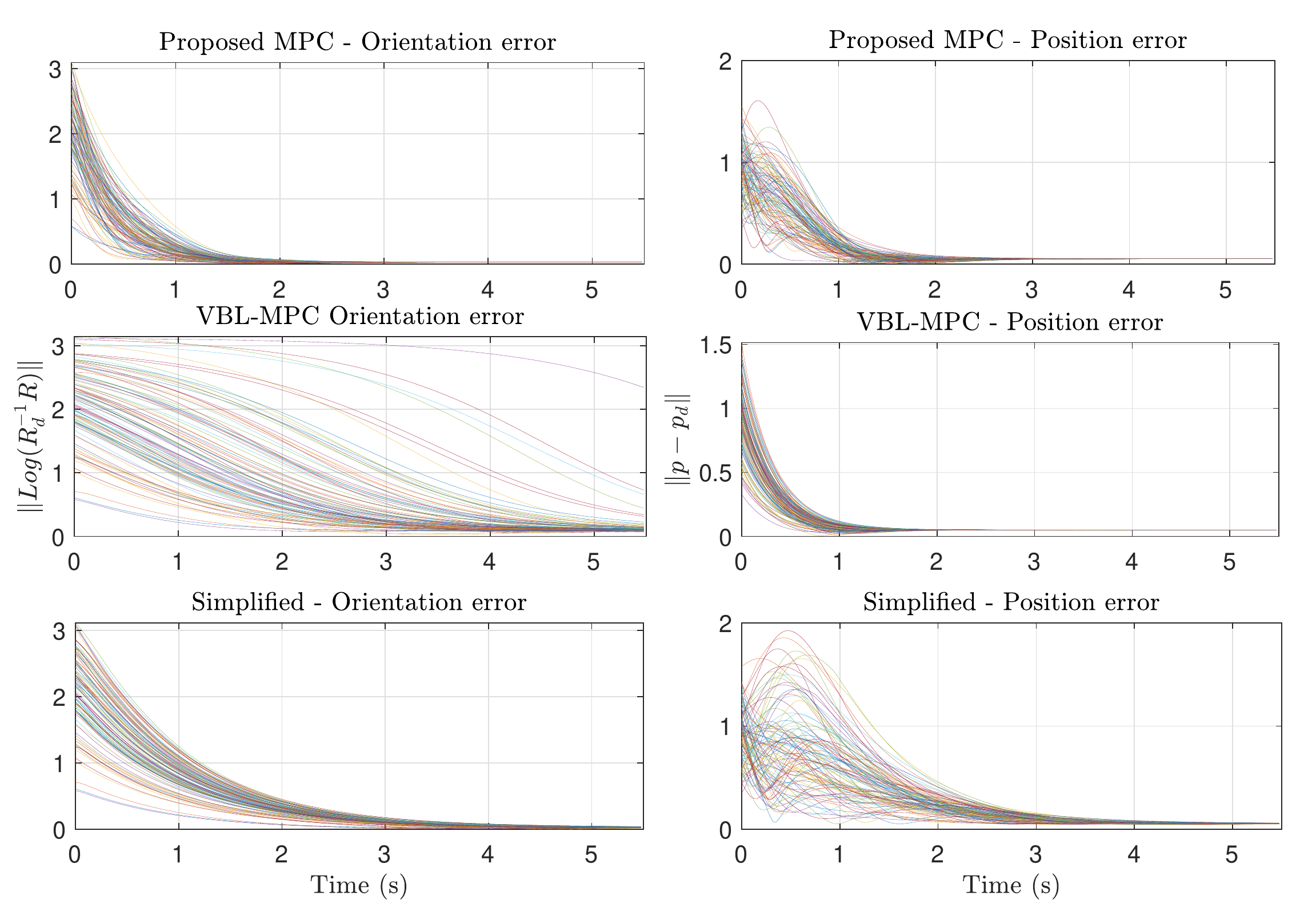}
    \caption{Tracking error of controllers in simulations with 100 randomly sampled initial poses. Our controller outperforms both baselines in the convergence rate of orientation error. The VBL-MPC deals with the position in $\mathbb{R}^3$, thus outperforming the other two methods. The simplified version of the proposed method is less accurate than the proposed method in both position and orientation tracking.}
    \label{fig:sim_tracking_error}
\end{figure}

In this section, we apply our controller on a fully actuated three-dimensional rigid body with the dynamics as shown in \eqref{eq:rigid_body_dynamics} for trajectory tracking. In this case, we do not consider the gravity and the system matrices are the same as we defined in Problem \ref{prob:cmpc}. The system inputs are the torque $\tau$ and force $f$ in the robot body frame $u:=\begin{bmatrix}
    \tau \\
    f
\end{bmatrix}$. We define a spiral curve with constant twists $$\xi_d = [0,0,1,2,0,0.2]^{\transpose}.$$ The reference trajectory is integrated by $\xi_d$  from the identity, i.e., $R_0 = I, p_0 = 0$. We randomly sample 100 initial orientations and positions around the identity and then apply ours and two baseline controllers. One baseline controller is the VBL-MPC proposed in \cite{vblMPC}. The VBL-based method uses the compatible error to parameterize the difference between two orientations as
\begin{equation}
    e_R := \frac{1}{2}(R^{-1}R_d - R_d^{-1}R)^{\vee} ,
\end{equation}
where $(\cdot)^\vee$ is the inverse of $(\cdot)^\wedge$ map. 
Another baseline is our method with a simplified matrix, which is the version of \eqref{eq:se3_error_lin} that does not consider the effect of the adjoint map in $A_t$ and $C_t$. The orientation part of this simplified version is the same as the local control law proposed in \cite{Ilya-SO3} and \cite{Ilya-SE3}. 

As the performance of MPC is strongly dependent on the parameter tuning, we keep the comparison fair by using the same stage quadratic cost $Q$ and $R$. The terminal cost is computed by the discrete-time Riccati equation
\begin{equation}
    P = A^{\transpose}PA - (A^{\transpose}PB)(R + B^{\transpose}PB)^{-1}(B^{\transpose}PA) + Q,
\end{equation}
which is intended to approximate the cost of the unconstrained problem in the infinite horizon with the pre-defined stage cost $Q$ and $R$.
All methods use the same control horizon $N = 12$ and the same constraints on the input. For longer horizons, there is no noticeable improvement in the tracking performance. 

\begin{figure}[t]
    \centering
    \includegraphics[width=1\columnwidth]{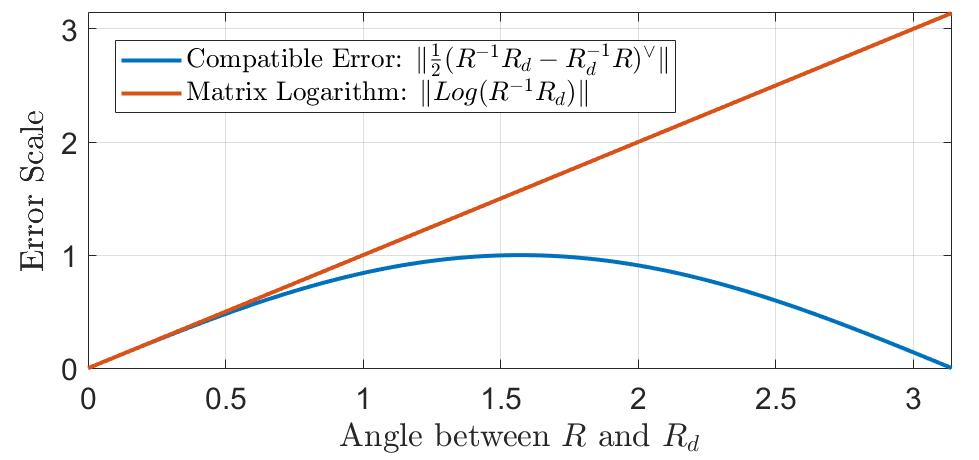}
    \caption{Comparison of different orientation errors. The compatible error for orientation does not reserve the scale of error. When the initial orientation error is large, the controller may not generate enough response, thus making the convergence slow.}
    \label{fig:error_compare}
\end{figure}

\begin{figure} 
    \centering
    \includegraphics[width=1\columnwidth]{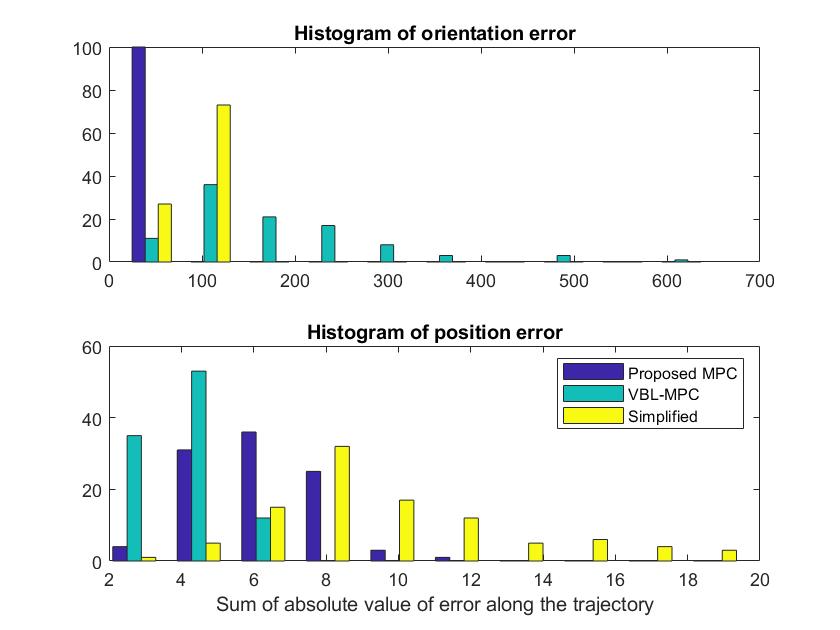}
    \caption{Histogram of tracking error of the 100 trials of simulation with randomly sampled initial poses. The horizontal axes indicate the sum of the error at the sample time along the trajectory presented in Fig.~\ref{fig:sim_tracking_error}. We can see that the orientation error of our controller remains small in most cases, while the VBL MPC has a long tail. The position tracking error remains small for the VBL MPC in most cases. Our method outperforms the simplified version. }
    \label{fig:sim_histogram}
\end{figure}

The reference trajectory and a sample trajectory of the proposed controller are presented in Fig.~\ref{fig:sim_spiral}. The tracking errors are presented in Fig.~\ref{fig:sim_tracking_error}. We can see that as the compatible error is used in VBL-based MPC, the convergence rate is much lower when the initial orientation error is large. This effect has also been shown in \cite{lee2010geometric}, where exponential stability is only guaranteed when the initial orientation error is less than $90{\degree}$. For a rough explanation of this effect, we plotted the scale of compatible error and the one obtained by matrix logarithm in Fig.~\ref{fig:error_compare}. When the orientation error approaches $180{\degree}$, the compatible error goes to $0$. A more theoretical explanation and proof can be found in~\cite{teng2022lie}. 

The position error of VBL-based MPC converges fast as it is decoupled from the orientation error. The histogram of the accumulated error along the simulated trajectory is presented in Fig.~\ref{fig:sim_histogram}. It is obvious that our controller outperforms both baselines in the orientation tracking. Our controller also outperforms the simplified version in both orientation and position tracking, which demonstrates the success of the linearization scheme. 

\section{Validation on quadrupedal robot}
We now conduct two experiments on the quadrupedal robot Mini Cheetah \cite{minicheetah} to evaluate the proposed MPC. Both experiments use a single rigid body model to approximate the torso motion. 
\subsection{Quadrupedal robot control}
The centroidal dynamics of the legged robot can be approximated by the rigid body equations of motion in~\eqref{eq:rigid_body_dynamics}. As the leg of Mini Cheetah is modeled as point contact with the ground, we assume only Ground Reaction Force (GRF) is acted on the contact point. We denote the GRF at the $k$-th leg as $f_{b,k}\in\mathbb{R}^3$. The torque acted on the center of mass is mapped from GRF by the lever arm of the legs. The vector from the robot center of mass to the $k$-th contact point is denoted as $r_{b,k} \in \mathbb{R}^3$. Note that the GRF  and lever arms are represented in the body frame; thus, we have the subscripts $b$. The friction cone constraints are considered in the world frame, denoted by subscript $w$ as 
\begin{equation}
\label{eq:fric_cons}
    f_w = Rf_b, |f_{w,x}| \le \mu f_{w,z}, |f_{w,y}| \le \mu f_{w,z}, f_{w,z}>0 .
\end{equation}
The lever arm and GRF are illustrated in Fig.~\ref{fig:first_fig}. 
Suppose the robot has $n$ legs on the ground. The continuous time error dynamics \eqref{eq:forcedEP_lin_2} can be represented by
\begin{equation}
\label{eq:dyn_dog}
    \dot{x}_t = A_tx_t + B_tu_t + h_t + \begin{bmatrix}
    0_{9\times 1}\\
    R_t^{\transpose}g
    \end{bmatrix},
\end{equation}
where $g$ denotes the gravity and $B_t$ becomes
\begin{equation}
\label{eq:robot_AtBt}
    B_t = \begin{bmatrix}
    0_3 & ... &0_3 \\
    0_3 & ... &0_3 \\
    I_b^{-1}r_{b,1,t}^{\wedge}& ... &I_b^{-1}r_{b,n,t}^{\wedge} \\
    \frac{{I}}{m} & ... &\frac{{I}}{m} \\
    \end{bmatrix}.
\end{equation}
To implement the convex QP algorithm, we assume that the lever arm $r_{b,k,t}$ remains constant during the planning horizon. We also assume the orientation $R$ in the gravity term $R^{\transpose}g$ in \eqref{eq:dyn_dog} and friction constraints \eqref{eq:fric_cons} remains constant. By zero-order hold or Euler first-order integration, we can obtain the discrete-time system matrix needed for Problem~\ref{prob:dtcmpc}.

We compare the proposed controller with two baseline controllers, the VBL-based MPC \cite{vblMPC} and the Euler angle-based MPC \cite{CheetahCMPC}. We do not use any feedforward term obtained by a high-level planner to make the comparison fair. The MPC stage cost and terminal cost settings are the same as the simulation. For real-time implementation, the terminal cost matrix $P$ is approximated by executing one step of Riccati recursion every time before the MPC is applied. By this method, $P$ will converge to the steady state value after a few iterations. For all these experiments, we choose the friction coefficient $\mu = 0.6$. 
\subsection{robot pose tracking}
\begin{figure}[t]
    \centering
    \includegraphics[width=0.9\columnwidth]{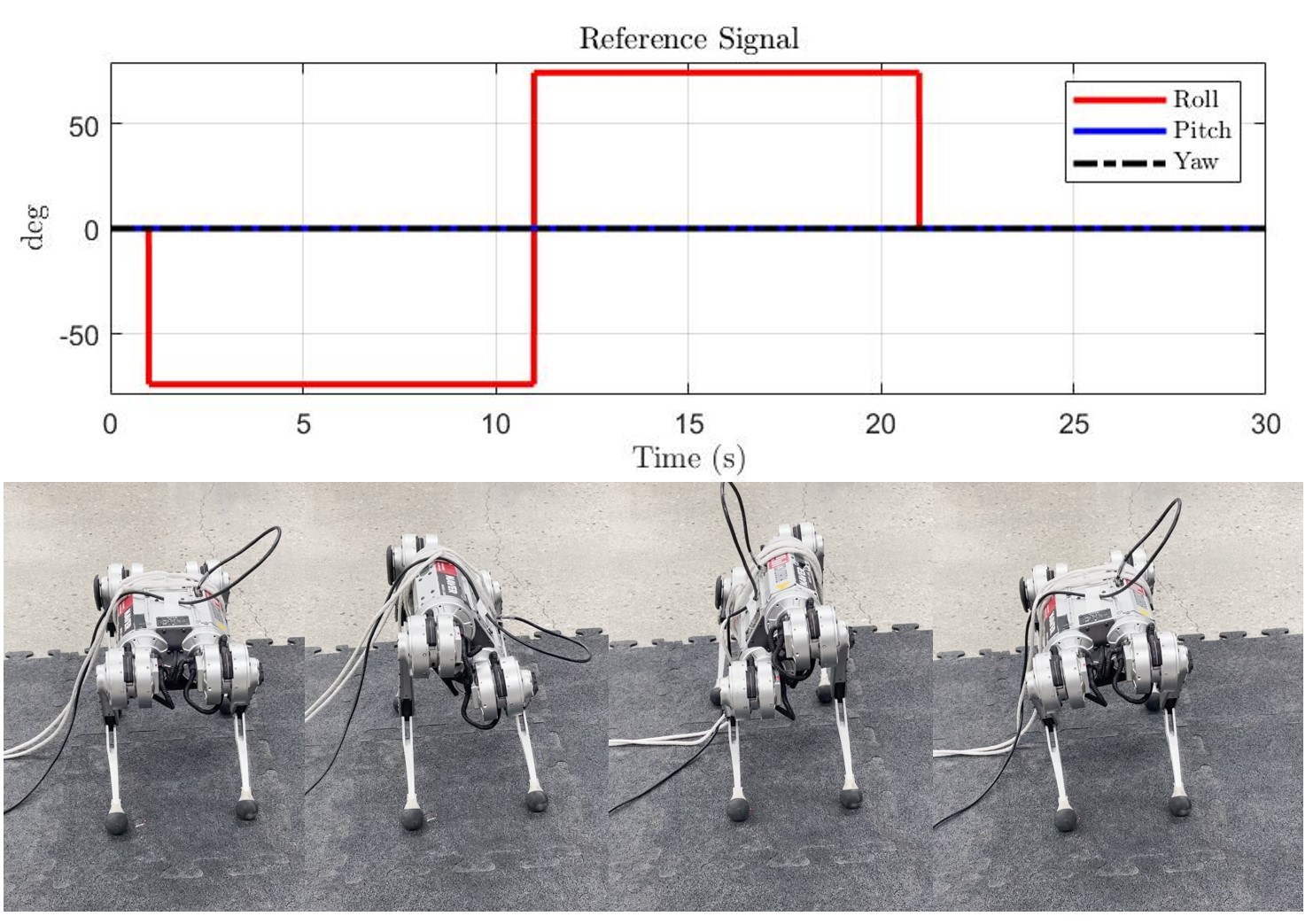}
    \caption{Reference signal for robot roll angle tracking. The robot roll angle changes from 0 to -74.5 degrees from 1 sec to 11 seconds. The yaw and pitch angle remains 0 along the reference trajectory. Then the robot leans to the opposite side for 10 seconds. }
    \label{fig:roll_signal_robot}
\end{figure}

\begin{figure}[t]
    \centering
    \includegraphics[width=0.9\columnwidth]{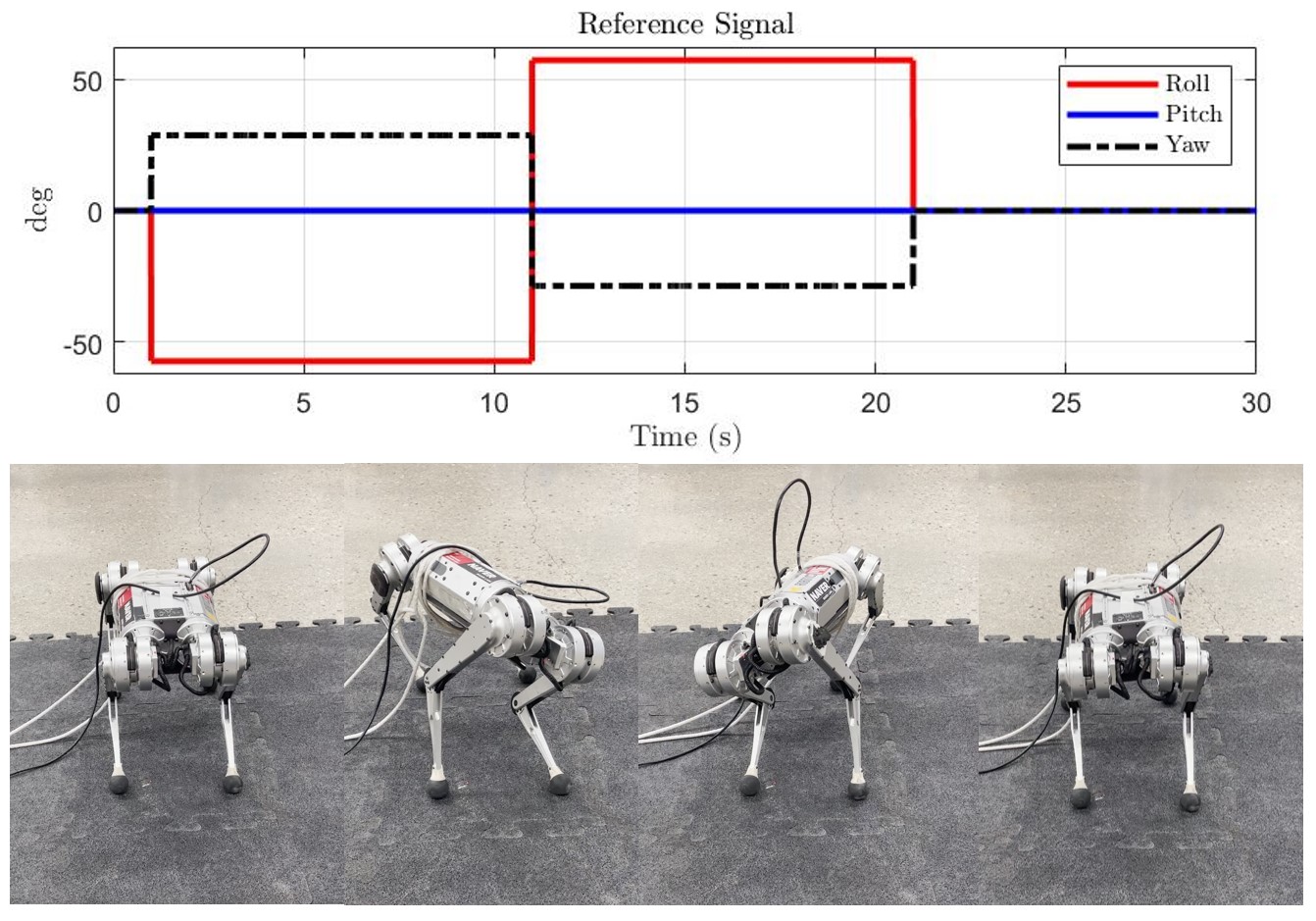}
    \caption{Reference signal for robot roll and yaw angle tracking. From 1 sec to 11 seconds, the robot roll changes from 0 to -57.3 degrees, and the yaw changes from 0 to 28.5 degrees. Then the robot leans to the opposite side for 10 seconds.}
    \label{fig:ry_signal_robot}
\end{figure}

In this experiment, we apply several step orientation signals for a robot to track. All four legs of robot are on the ground so the $B_t$ matrix in \eqref{eq:robot_AtBt} has four blocks. Step signals of pure robot roll angle and mixture of roll and yaw angle are applied. The GRF planned by the MPC is mapped to the joint torques $\tau_{st}$ by the spatial Jacobian $J$ via
\begin{equation}
    \tau_{st} = -J^{\transpose}f_w .
\end{equation}
The reference signals and snapshots of robot motion are presented in Fig.~\ref{fig:roll_signal_robot} and \ref{fig:ry_signal_robot}. As terminal cost is well designed, we use a small control horizon $N=4$ and $\Delta t = 0.025 s$ in all the experiments. Each experiment is conducted three times to eliminate the influence of random factors. 

\begin{figure*}
    \centering
    \includegraphics[width=2\columnwidth]{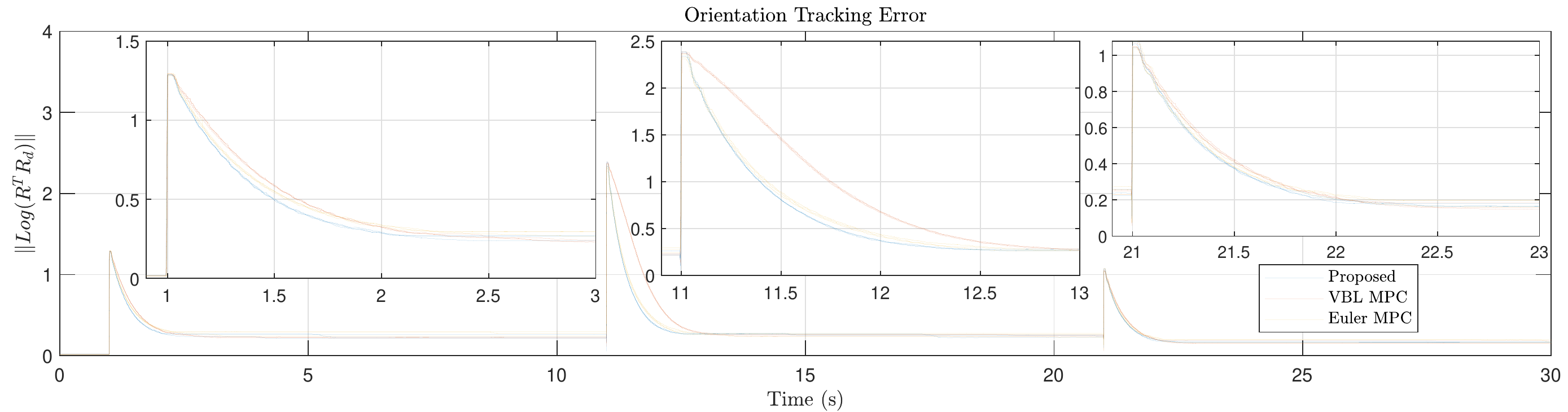}
    \caption{Error convergence for roll tracking. When a new step signal is applied, our controller converges faster than the baseline methods and exhibits smaller steady-state error. The VBL-MPC is slower due to the use of the compatible error. As only roll signal is applied, the errors defined on SE(3) and Euler are the same. Thus, the Proposed MPC and Euler MPC have similar tracking performance.}
    \label{fig:roll_error}
\end{figure*}

\begin{figure*}
    \centering
    \includegraphics[width=2\columnwidth]{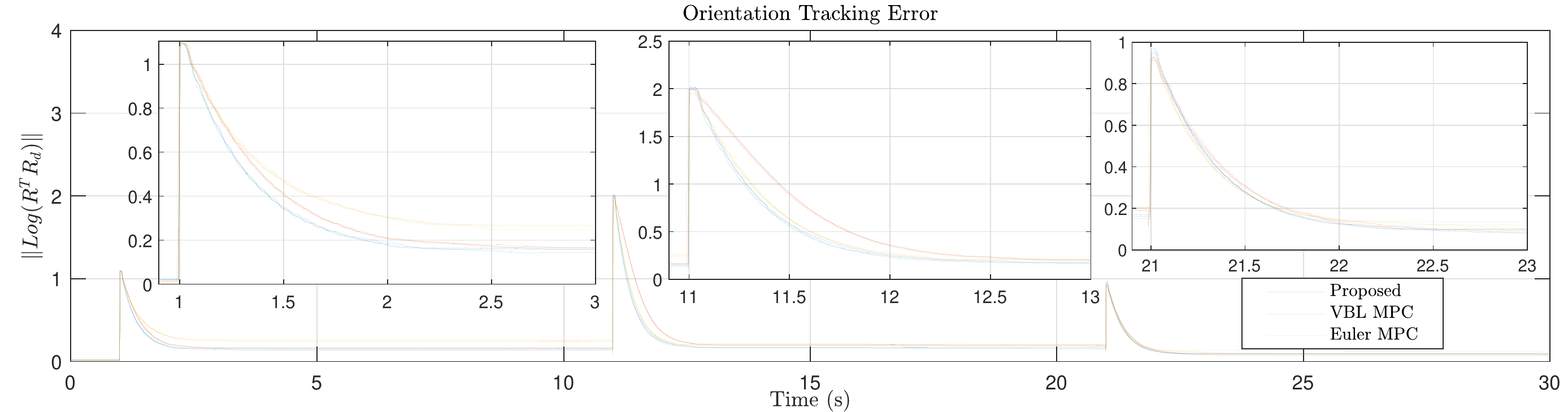} %
    \caption{Error convergence for roll and yaw tracking. When a new step signal is applied, our controller converges faster than the baseline methods and has a smaller steady-state error. The Euler angle-based MPC has a larger steady-state error as both roll and yaw signals are applied.}
    \label{fig:ry_error}
\end{figure*}

The details of the responses are presented in Fig.~\ref{fig:roll_error} and \ref{fig:ry_error}. It can be seen that as no feedforward force at the equilibrium is provided, all three controllers have steady-state error. However, the geometric-based controller, i.e., proposed and the VBL-based MPC, has a smaller steady-state error than the Euler angle-based one. As the VBL-based MPC does not conserve the scale of the error, the convergence rate is much lower than our controller, which is obvious when the opposite Euler angle signal is applied at the middle of the reference profile. The convergence rate is consistent with the numerical simulation.

\subsection{Robot trotting}
We also apply our controller to robot locomotion. All controllers are based on the open-source software developed by \cite{wbic}. We use the parameters provided in the original MPC. The control horizon and discrete timestep are set to be $N=10$ and $ \Delta t = 0.0.25s$. This control horizon is the shortest one that ensure stable walking gaits. Ours and baseline controllers are deployed to plan the robot's GRF given command twists. Then the GRF is applied to the Whole Body Impulse Control (WBIC)~\cite{wbic} to obtain the joint torques. The WBIC decomposes the cartesian space task to joint space according to different hierarchies via the Jacobian null space decomposition, making the joint space motion much smoother than direct PD tracking. Unlike the conventional whole-body controller, WBIC prioritizes the GRF generation by penalizing the deviation of GRF from the planned GRF. We increase this penalty by 1e4 times in the original WBIC, so the GRF merely deviates from the planned one. 

\begin{figure}[t]
    \centering
    \includegraphics[width=1\columnwidth]{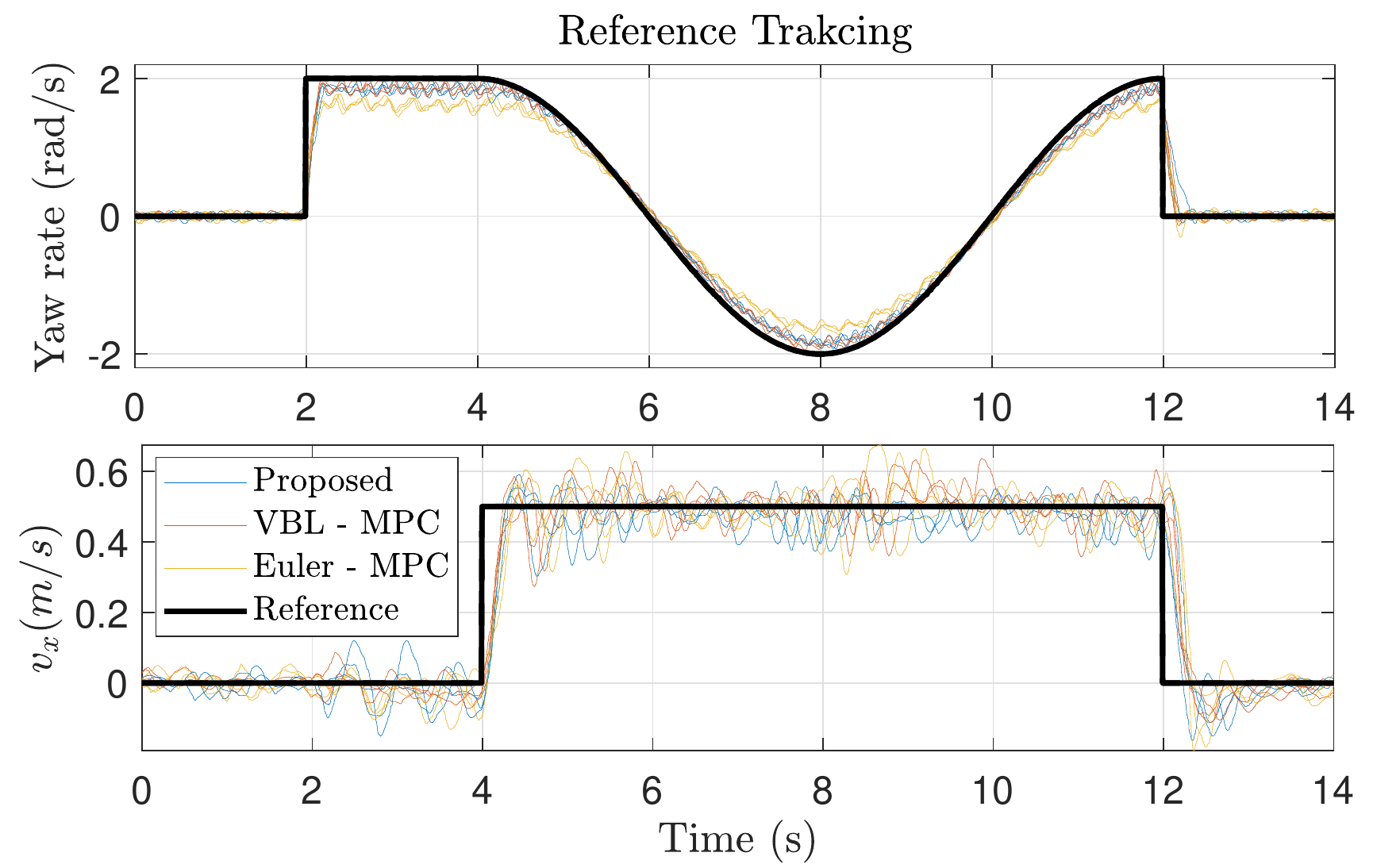}
    \caption{Reference tracking for quadrupedal robot trotting. Each controller is tested three times. The responses are too noisy; thus, the results are smoothed using the moving average filter.} 
    \label{fig:walking_track}
\end{figure}

\begin{figure}[t]
    \centering
    \includegraphics[width=1\columnwidth]{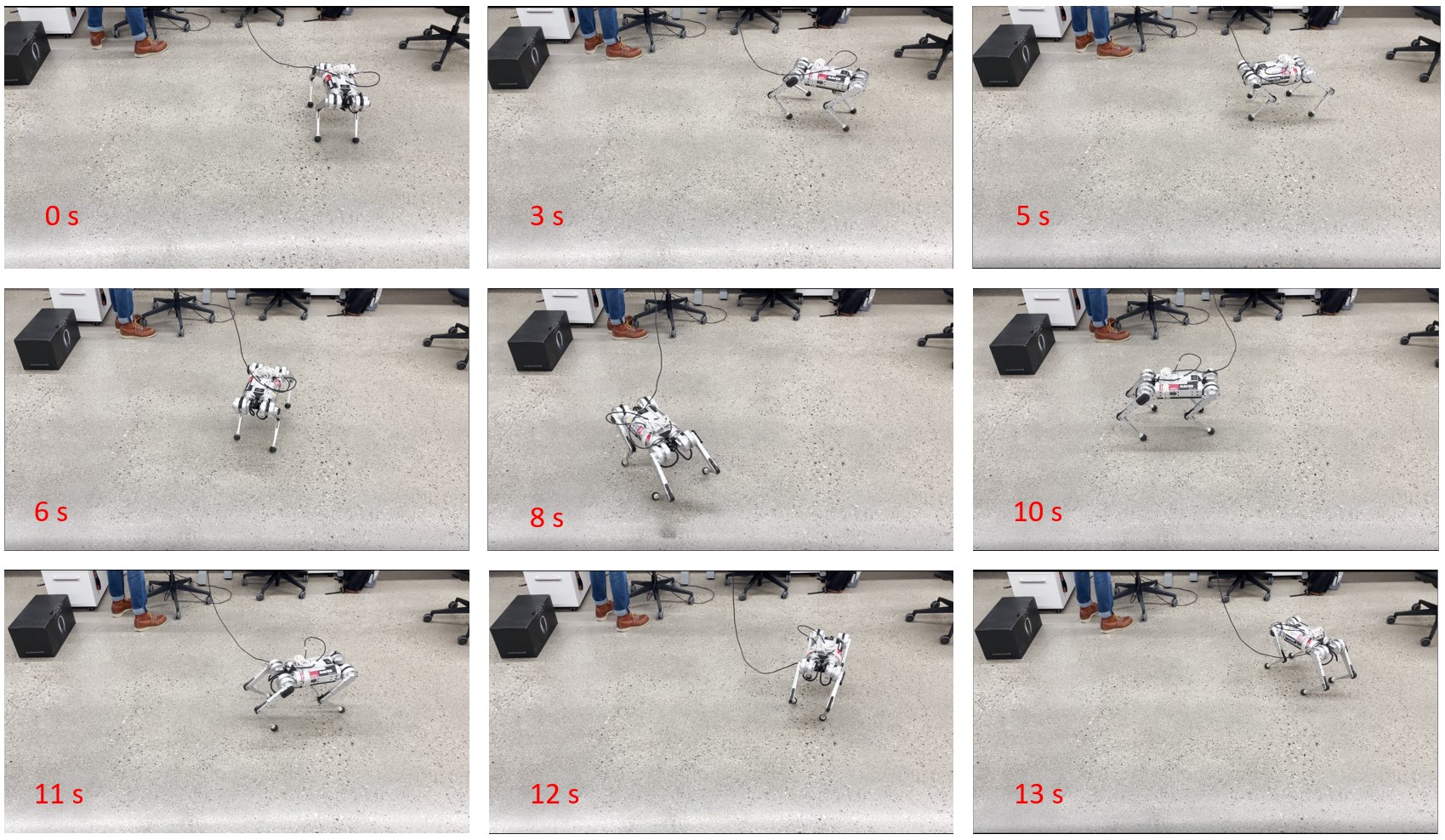}
    \caption{Snapshots of the experiments on reference tracking in Mini Cheetah trotting. The time corresponds to the reference signal in Fig.~\ref{fig:walking_track}.} 
    \label{fig:walking_exp}
\end{figure}

We first apply a step signal in the yaw rate. Then we add a step signal in forward motion in the robot frame, and the yaw rate becomes a sinusoidal signal. The reference is presented in Fig.~\ref{fig:walking_track} and the snapshots of the experiments are in Fig.~\ref{fig:walking_exp}. The tracking result is shown in Fig.~\ref{fig:walking_track}. We find that ours and the VBL-MPC can better track the yaw rate than the Euler angles-based MPC, as expected. All controllers can track the linear velocity well. The two baselines deal with the linear velocity in $\mathbb{R}^3$ space. As every step the orientation and position tracking errors are integrated from the current state, it is reasonable that all controllers perform well.

\section{Discussions}
In the robot pose tracking experiments, the large steady-state error is probably due to the friction of the mechanical part. In the simulation environments of Mini Cheetah, we noticed that the steady-state error is much smaller than observed in the experiments. The Euler angle-based MPC has larger steady-state errors than geometric-based ones, which we believe is due to the loss of symmetry. The Euler angle defines the rotation with respect to fixed axes, while the torque generates simultaneous rotation around body axes.

We showed the advantage of the proposed MPC over the variational-based MPC in terms of the convergence rate of orientation error. This advantage is attributed to the quadratic cost function designed in the Lie algebra. We compared the difference intuitively by depicting scales of the compatible and logarithm error. A preliminary result of the exponential convergence rate of linear feedback controllers on Lie groups by constructing the Lyapunov function in the Lie algebra has been discussed in \cite{teng2022lie}. 

We derived the linearized dynamics in continuous time and used the Euler first-order integration for implementation. An integration scheme that preserves the Lagrangian can be integrated with the proposed framework in future work.

\section{Conclusions}
We developed a new error-state Model Predictive Control approach on connected matrix Lie groups for robot control. By exploiting the existing symmetry of the pose control problem on $\mathrm{SE}(3)$ Lie group, we showed that the linearized tracking error dynamics and equations of motion in the Lie algebra are globally valid and evolve independently of the system trajectory. In addition, we formulated a convex MPC program for solving the problem efficiently using QP solvers. The simulation and experimental results confirm that the proposed approach provides faster convergence when rotation and position are controlled simultaneously. 

Future work includes the extension of the developed controller with learning-aided state estimators~\cite{lin2021legged} to enable environmental awareness and more aggressive maneuvers. 

{\small 
\balance
\bibliographystyle{IEEEtran}
\bibliography{bib/strings-abrv,bib/ieee-abrv,bib/references}
}

\end{document}